\documentclass[11pt,a4paper]{article}
\usepackage{chapterbib}
\usepackage[hyperref]{naaclhlt2019}
\usepackage{times}
\usepackage{latexsym}

\usepackage{graphicx}
\usepackage{tabularx}
\usepackage{balance}
\usepackage{xcolor}
\usepackage{subcaption}
\usepackage{multirow}
\usepackage{titling}

\usepackage{amsmath}
\usepackage{breqn}
\usepackage{xpatch}

\usepackage{url}

\newcommand\blfootnote[1]{%
	\begingroup
	\renewcommand\thefootnote{}\footnote{#1}%
	\addtocounter{footnote}{-1}%
	\endgroup
}

\begin{document}

	%
%
%

\aclfinalcopy 
\def\aclpaperid{514} 


\newcommand\BibTeX{B{\sc ib}\TeX}

\definecolor{limeGreen}{RGB}{196, 214, 160}
\definecolor{normalGreen}{RGB}{141, 208, 63}
\definecolor{lightBlue}{RGB}{141, 177, 226}
\definecolor{whitewhite}{RGB}{255,255,255}

\title{\Large{\bf{ReWE: Regressing Word Embeddings\\for Regularization of Neural Machine Translation Systems}}}

\author{\bf{Inigo Jauregi Unanue\textsuperscript{1,2}, Ehsan Zare Borzeshi\textsuperscript{$*$,2}, Nazanin Esmaili\textsuperscript{2}, Massimo Piccardi\textsuperscript{1}} \\
\textsuperscript{1} University of Technology Sydney, Sydney, Australia \\
\textsuperscript{2} Capital Markets Cooperative Research Centre, Sydney, Australia \\
\\
{\tt \{ijauregi,ezborzeshi,nesmaili\}@cmcrc.com}\\ 
{\tt massimo.piccardi@uts.edu.au}
}

\date{}

\maketitle
\thispagestyle{empty}
\begin{abstract}
Regularization of neural machine translation is still a significant problem, especially in low-resource settings. To mollify this problem, we propose regressing word embeddings (ReWE) as a new regularization technique in a system that is jointly trained to predict the next word in the translation (categorical value) and its word embedding (continuous value). Such a joint training allows the proposed system to learn the distributional properties represented by the word embeddings, empirically improving the generalization to unseen sentences. Experiments over three translation datasets have showed a consistent improvement over a strong baseline, ranging between $0.91$ and $2.54$ BLEU points, and also a marked improvement over a state-of-the-art system.
\end{abstract}

\section{Introduction}
\label{sec:Introduction}

The last few years \blfootnote{\textsuperscript{$*$} The author has changed affiliation to Microsoft after the completion of this work. His new email is: \tt Ehsan.ZareBorzeshi@microsoft.com} have witnessed remarkable improvements in the performance of machine translation (MT) systems. These improvements are strongly linked to the development of neural machine translation (NMT): based on encoder-decoder architectures (also known as seq2seq), NMT can use recurrent neural networks (RNNs) \cite{sutskever2014,cho2014,wu2016google}, convolutional neural networks (CNNs) \cite{gehring2016convolutional} or transformers \cite{vaswani2017attention} to learn how to map a sentence from the source language to an adequate translation in the target language. In addition, attention mechanisms \cite{bahdanau2014,luong2015} help soft-align the encoded source words with the predictions, further improving the translation.

NMT systems are usually trained via maximum likelihood estimation (MLE). However, as pointed out by \cite{elbayad2018token}, MLE suffers from two obvious limitations: the first is that it treats all the predictions other than the ground truth as equally incorrect. As a consequence, synonyms and semantically-similar words --- which are often regarded as highly interchangeable with the ground truth --- are completely ignored during training. The second limitation is that MLE-trained systems suffer from ``exposure bias'' \cite{bengio2015scheduled,ranzato2015sequence} and do not generalize well over the large output space of translations. Owing to these limitations, NMT systems still struggle to outperform other traditional MT approaches when the amount of supervised data is limited \cite{koehn2017six}. 

\begin{figure}[t!]
	\centering
	\includegraphics[width=0.9\linewidth]{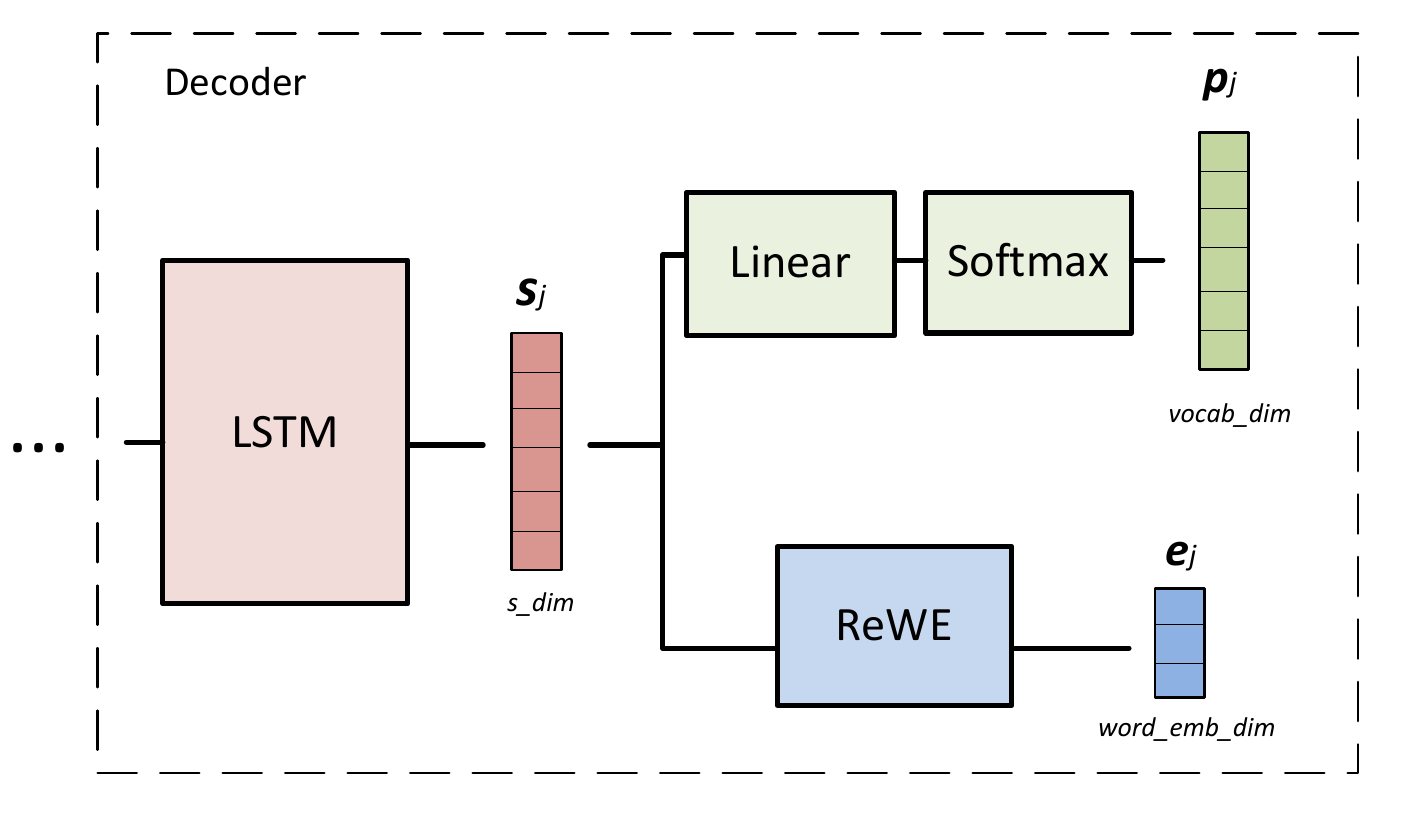}
	\caption{The proposed regularizer: the hidden vector in the decoder, $s_j$, transits through two paths: 1) a linear and a softmax layers that output vector $v_j$ (vocab\_dim) which is used for predicting the target word as usual, and 2) a two-layer network (ReWE) that outputs a vector, $e_j$, of word embedding size (word\_emb\_dim). During training, $e_j$ is used in a regressive loss with the ground-truth embedding.}
	\label{fig:fig1}
	\vspace{-12pt}
\end{figure}

In this paper, we propose a novel regularization technique for NMT aimed to influence model learning with contextual properties. The technique --- nicknamed ReWE from ``regressing word embedding'' --- consists of modifying a conventional seq2seq decoder to jointly learn to a) predict the next word in the translation (categorical value), as usual, and b) regress its word embedding (numerical value). Figure \ref{fig:fig1} shows the modified decoder. Both predictions are incorporated in the training objective, combining standard MLE with a continuous loss function based on word embeddings. The rationale is to encourage the system to learn to co-predict the next word together with its context (by means of the word embedding representation), in the hope of achieving improved generalization. At inference time, the system operates as a standard NMT system, retaining the categorical prediction and ignoring the predicted embedding. We qualify our proposal as a regularization technique since, like any other regularizers, it only aims to influence the model's training, while leaving the inference unchanged. We have evaluated the proposed system over three translation datasets of different size, namely English-French (en-fr), Czech-English (cs-en), and Basque-English (eu-en). In each case, ReWE has significantly outperformed its baseline, with a marked improvement of up to 2.54 BLEU points for eu-en, and consistently outperformed a state-of-the-art system~\cite{denkowski2017stronger}. 

\section{Related work}
\label{sec:Related_work}

A substantial literature has been devoted to improving the generalization of NMT systems. Fadaee et al. \shortcite{fadaee2017data} have proposed a data augmentation approach for low-resource settings that generates synthetic sentence pairs by replacing words in the original training sentences with rare words. Kudo \shortcite{kudo2018subword} has trained an NMT model with different subword segmentations to enhance its robustness, achieving consistent improvements over low-resource and out-of-domain settings. Zhang et al. \shortcite{zhang2018regularizing} have presented a novel regularization method that encourages target-bidirectional agreement.
Other work has proposed improvements over the use of a single ground truth for training: Ma et al. \shortcite{ma2018bag} have augmented the conventional seq2seq model with a bag-of-words loss under the assumption that the space of correct translations share similar bag-of-words vectors, achieving promising results on a Chinese-English translation dataset; Elbayad et al. \shortcite{elbayad2018token} have used sentence-level and token-level reward distributions to ``smooth'' the single ground truth. Chousa et al. \shortcite{chousa2018training} have similarly leveraged a token-level smoother.

In a recent paper, Denkowski and Neubig \shortcite{denkowski2017stronger} have achieved state-of-the-art translation accuracy by leveraging a variety of techniques which include: dropout \cite{srivastava2014dropout}, lexicon bias \cite{arthur2016incorporating}, pre-translation \cite{niehues2016pre}, data bootstrapping \cite{chen2016guided}, byte-pair encoding \cite{sennrich2016neura} and ensembles of independent models \cite{rokach2010}.

However, to our knowledge none of the mentioned approaches have explicitly attempted to leverage the embeddings of the ground-truth tokens as targets. For this reason, in this paper we explore regressing toward pre-trained word embeddings as an attempt to capture contextual properties and achieve improved model regularization.


\section{Model}
\label{sec:model}

\subsection{Seq2seq baseline}
\label{sub_sec:Seq2seq_baseline}

The model is a standard NMT model with attention in which we use RNNs for the encoder and decoder. Following the notation of \cite{bahdanau2014}, the RNN in the decoder generates a sequence of hidden vectors, $\{\textbf{s}_1,\dots,\textbf{s}_m\}$, given the context vector, the previous hidden state $\textbf{s}_{j-1}$ and the previous predicted word $\textbf{y}_{j-1}$:

\vspace{-12pt}

\begin{equation}
\label{eq:decoder}
\begin{split}
\textbf{s}_{j}=dec_{rnn}(\textbf{s}_{j-1},\textbf{y}_{j-1},\textbf{c}_j) \quad j=1,\dots,m 
\end{split}
\end{equation}

\noindent where $y_0$ and $s_0$ are initializations for the state and label chains. Each hidden vector $\textbf{s}_j$ (of parameter size $S$) is then linearly transformed into a vector of vocabulary size, $V$, and a softmax layer converts it into a vector of probabilities (Eq. \ref{eq:generator}), where $W$ (a matrix of size $V \times S$) and $b$ (a vector of size $V \times 1$) are learnable parameters. The predicted conditional probability distribution over the words in the target vocabulary, $\textbf{p}_j$, is given as:

\vspace{-12pt}

\begin{equation}
\label{eq:generator}
\begin{split}
\textbf{p}_{j}= softmax(\textbf{W}\textbf{s}_j+\textbf{b}) 
\end{split}
\end{equation}

As usual, training attempts to minimize the negative log-likelihood (NLL), defined as:

\begin{equation}
\label{eq:NLL_loss}
\begin{split}
NLL_{loss} = -\sum_{j=1}^{m}\log(\textbf{p}_{j}(\textbf{y}_{j}))
\end{split}
\end{equation}

\noindent where $\textbf{p}_{j}(\textbf{y}_{j})$ notes the probability of ground-truth word $\textbf{y}_j$. The NLL loss is minimized when the probability of the ground truth is one and that of all other words is zero, treating all predictions different from the ground truth as equally incorrect.

\subsection{ReWE}
\label{sub_sec:Combining_cat_and_num_losses}


Pre-trained word embeddings \cite{pennington2014glove,bojanowski2016enriching,mikolov2013distributed} capture the contextual similarities of words, typically by maximizing the probability of word $w_{t+k}$ to occur in the context of center word $w_t$. This probability can be expressed as:
\begin{equation}
\label{eq:word_embedding}
\begin{split}
p(w_{t+k}|w_t),  \quad &-c \leq k \leq c, k \neq 0 \\
&\quad t=1,\dots,T 
\end{split}
\end{equation} 
\noindent where $c$ is the size of the context and $T$ is the total number of words in the training set. Traditionally, word embeddings have only been used as input representations. In this paper, we instead propose using them in output as part of the training objective, in the hope of achieving regularization and improving prediction accuracy. Building upon the baseline model presented in Section \ref{sub_sec:Seq2seq_baseline}, we have designed a new ``joint learning'' setting: our decoder still predicts the probability distribution over the vocabulary, $\textbf{p}_j$ (Eq. \ref{eq:generator}), while simultaneously regressing the same shared $\textbf{s}_j$ to the ground-truth word embedding, $e(\textbf{y}_j)$. The ReWE module consists of two linear layers with a Rectified Linear Unit (ReLU) in between, outputting a vector $\textbf{e}_j$ of word embedding size (Eq. \ref{eq:emb_generator}). Please note that adding this extra module adds negligible computational costs and training time. Full details of this module are given in the supplementary material.

\vspace{-6pt} 
\begin{equation}
\label{eq:emb_generator}
\begin{split}
\textbf{e}_j &= ReWE(\textbf{s}_j) \\ &= \textbf{W}_2(ReLU(\textbf{W}_1\textbf{s}_j+\textbf{b}_1))+\textbf{b}_2
\end{split}
\end{equation}


The training objective is a numerical loss, $l$ (Eq. \ref{eq:cont_loss}), computed between the output vector, $\textbf{e}_j$, and the ground-truth embedding, $e(\textbf{y}_j)$: 

\vspace{-15pt}

\begin{equation}
\label{eq:cont_loss}
\begin{split}
ReWE_{loss} = l(\textbf{e}_j,e(\textbf{y}_j))
\end{split}
\end{equation}

\vspace{-3pt}

In the experiment, we have explored two cases for the $ReWE_{loss}$: the minimum square error (MSE)\footnote{https://pytorch.org/docs/stable/nn.html\#torch.nn.\\MSELoss} and the cosine embedding loss (CEL)\footnote{https://pytorch.org/docs/stable/nn.html\#torch.nn.\\CosineEmbeddingLoss}. Finally, the $NLL_{loss}$ and the $ReWE_{loss}$ are combined to form the training objective using a positive trade-off coefficient, $\lambda$:

\vspace{-20pt}

\begin{equation}
\label{eq:combined_loss}
\begin{split}
Loss = NLL_{loss}+\lambda ReWE_{loss}
\end{split}
\end{equation}

\vspace{-3pt}

As mentioned in the Introduction, at inference time we ignore the ReWE output, $\textbf{e}_j$, and the model operates as a standard NMT system. 

\section{Experiments}
\label{sec:Experiments}

We have developed our models building upon the OpenNMT toolkit \cite{klein2017opennm}\footnote{Our code can be found at: https://github.com/ijauregiCMCRC/ReWE\_NMT}. For training, we have used the same settings as \cite{denkowski2017stronger}. We have also explored the use of sub-word units learned with byte pair encoding (BPE)~\cite{sennrich2016neura}. All the preprocessing steps, hyperparameter values and training parameters are described in detail in the supplementary material to ease reproducibility of our results.

We have evaluated these systems over three publicly-available datasets from the 2016 ACL Conference on Machine Translation (WMT16)\footnote{WMT16: http://www.statmt.org/wmt16/} and the 2016 International Workshop on Spoken Language Translation (IWSLT16)\footnote{IWSLT16: https://workshop2016.iwslt.org/}. 
Table \ref{tab:3} lists the datasets and their main features. Despite having nearly 90,000 parallel sentences, the eu-en dataset only contains 2,000 human-translated sentences; the others are translations of Wikipedia page titles and localization files. Therefore, we regard the eu-en dataset as very low-resource.

\begin{table}[t]
	\centering
	\resizebox{0.4\textwidth}{!}{\begin{tabularx}{1.2\columnwidth}{|l|l|X|}
			
			\hline
			\textbf{Dataset}&\textbf{Size}&\textbf{Sources}\\
			\hline
			IWSLT16 en-fr&$219,777$&TED talks\\
			IWSLT16 cs-en&$114,243$&TED talks\\
			\hline
			WMT16 eu-en&$89,413$&IT-domain data\\
			\hline
	\end{tabularx}}
	\label{tab:2}
	

\vspace{6pt}

	\centering
	\resizebox{0.4\textwidth}{!}{\begin{tabularx}{1.2\columnwidth}{|l|l|X|}
			
			\hline
			\textbf{Dataset}&\textbf{Validation set}&\textbf{Test set}\\
			\hline
			en-fr&TED test 2013+2014&TED test 2015+2016\\
			cs-en&TED test 2012+2013&TED test 2015+2016\\
			\hline
			eu-en&Sub-sample of PaCo&IT-domain test\\
			\hline
	\end{tabularx}}
	
	\caption{Top: parallel training data. Bottom: validation and test sets.}\label{tab:3}
	
	\vspace{-20pt}
	
\end{table}


In addition to the seq2seq baseline, we have compared our results with those recently reported by Denkowski and Neubig for non-ensemble models~\shortcite{denkowski2017stronger}.
For all models, we report the BLEU scores~\cite{papineni2002bleus}, with the addition of selected comparative examples. Two contrastive experiments are also added in supplementary notes.




\begin{figure}[t!]
	\centering
	\includegraphics[width=0.95\linewidth]{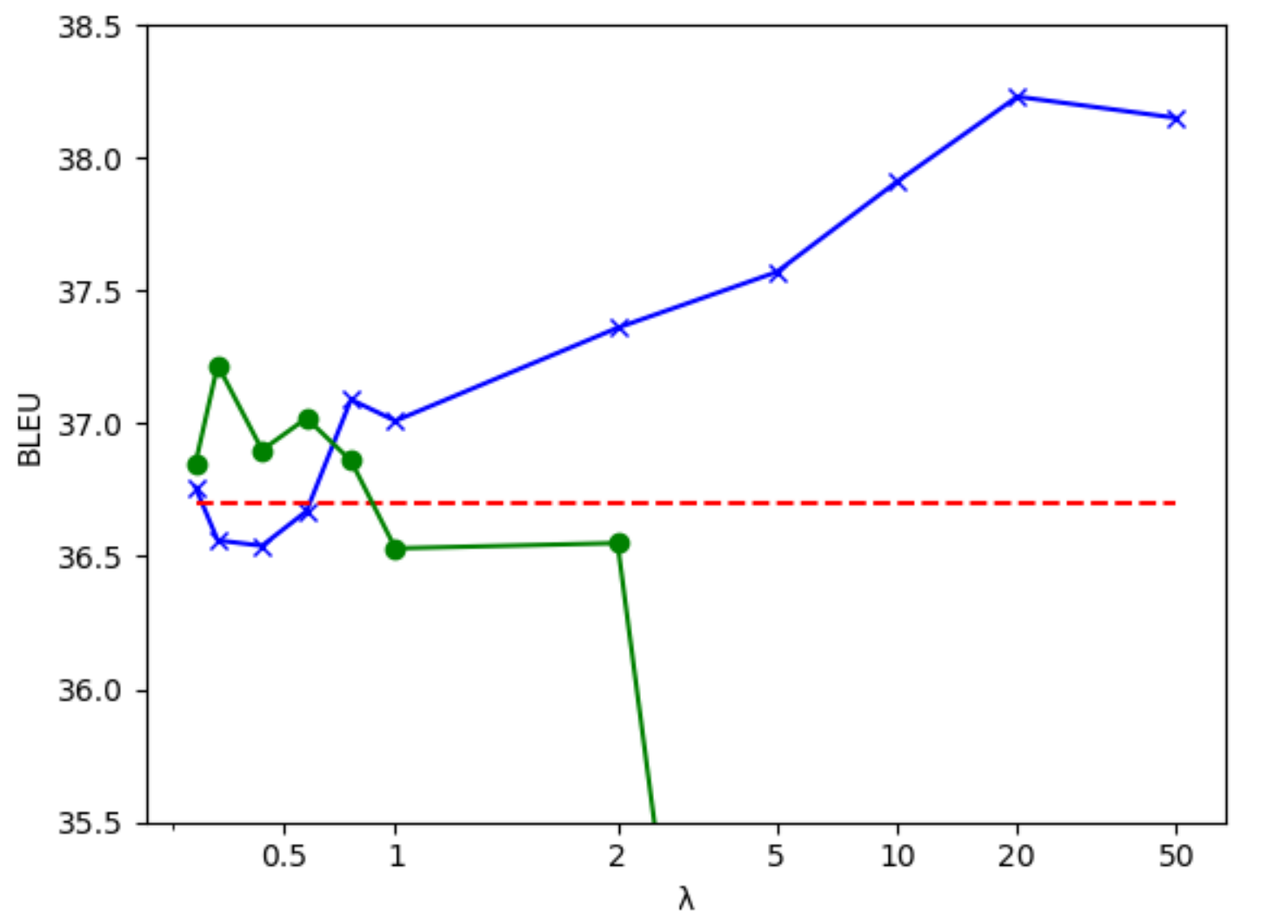}
	\caption{BLEU scores of three models over the en-fr validation set for different $\lambda$ values: baseline (\textcolor{red}{\textbf{red}}), baseline + ReWE (MSE) (\textcolor{green}{\textbf{green}}), baseline + ReWE (CEL) (\textcolor{blue}{\textbf{blue}}). Each point in the graph is an average of 3 independently trained models.}
	\label{fig:lambda}
	\vspace{-8pt}
\end{figure}


\begin{table*}[t]
	\centering
	\resizebox{0.8\textwidth}{!}{\begin{tabularx}{2.05\columnwidth}{|l|l l|l l|l X|}
			
			\hline
			\textbf{Models}&\multicolumn{2}{c|}{\textbf{en-fr}}&\multicolumn{2}{c|}{\textbf{cs-en}}&\multicolumn{2}{c|}{\textbf{eu-en}}\\
			&Word&BPE&Word&BPE&Word&BPE\\
			\hline
			\cite{denkowski2017stronger} &33.60&34.50&21.00&22.60&&\\
            \cite{denkowski2017stronger} + Dropout &34.5&34.70&21.4&23.60&&\\
            \cite{denkowski2017stronger} + Lexicon &33.9&34.80&20.6&22.70&&\\
            \cite{denkowski2017stronger} + Pre-translation &N/A&34.90&N/A&\textbf{23.80}&&\\
            \cite{denkowski2017stronger} + Bootstrapping &34.40&35.20&21.60&23.60&&\\
            \hline
            Our baseline &34.16&34.09&20.57&22.69&12.14&17.17\\
            Our baseline + ReWE (CEL) ($\lambda=20$) &\textbf{35.52}&\textbf{35.22}&\textbf{21.83}&23.60&\textbf{13.73}&\textbf{19.71}\\
\hline
	\end{tabularx}}
\caption{BLEU scores over the test sets. Average of 10 models independently trained with different seeds.}\label{tab:results}
    \vspace{-6pt}	
\end{table*}

\subsection{Results}
\label{sub_sec:results}

As a preliminary experiment, we have carried out a sensitivity analysis to determine the optimal value of the trade-off coefficient, $\lambda$ (Eq. \ref{eq:cont_loss}), using the en-fr validation set. The results are shown in Figure \ref{fig:lambda}, where each point is the average of three runs trained with different seeds. The figure shows that the MSE loss has outperformed slightly the baseline for small values of $\lambda$ ($< 1$), but the BLEU score has dropped drastically for larger values. Conversely, the CEL loss has increased steadily with $\lambda$, reaching $38.23$ BLEU points for $\lambda=20$, with a marked improvement of $1.53$ points over the baseline. This result has been encouraging and therefore for the rest of the experiments we have used CEL as the $ReWE_{loss}$ and kept the value of $\lambda$ to $20$. In Section \ref{sub_sec:discussion}, we further discuss the behavior of CEL and MSE.

Table \ref{tab:results} reports the results of the main experiment for all datasets. The values of our experiments are for blind runs over the test sets, averaged over $10$ independent runs with different seeds. The results show that adding ReWE has significantly improved the baseline in all cases, with an average of $1.46$ BLEU points. In the case of the eu-en dataset, the improvement has reached $2.54$ BLEU points. We have also run unpaired t-tests between our baseline and ReWE, and the differences have proved statistically significant ($p$-values $< 0.05$) in all cases. Using BPE has proved beneficial for the cs-en and eu-en pairs, but not for the en-fr pair. We speculate that English and French may be closer to each other at word level and, therefore, less likely to benefit from the use of sub-word units. Conversely, Czech and Basque are morphologically very rich, justifying the improvements with BPE. 

Table \ref{tab:results} also shows that our model has outperformed almost all the state-of-the-art results reported in \cite{denkowski2017stronger} (dropout, lexicon bias, pre-translation, and bootstrapping), with the only exception of the pre-translation case for the cs-en pair with BPE. This shows that the proposed model is competitive with contemporary NMT techniques.

\begin{table}[t]
	\centering
	\resizebox{0.45\textwidth}{!}{\begin{tabularx}{1.3\columnwidth}{|l X|}
			\hline
			\textbf{Src}: &Hautatu Kontrol panela $\rightarrow$ Programa lehenetsiak , eta aldatu bertan .\\
			\hline
			\textbf{Ref}: &Go to Control Panel $\rightarrow$ \textcolor{normalGreen}{Default programs} , and change it there .\\
			\hline
			\textbf{Baseline}: &Select the Control Panel $\rightarrow$ \textcolor{red}{program} , and change .\\
			\hline
			\textbf{Baseline + ReWE}: &Select the Control Panel $\rightarrow$ \textcolor{normalGreen}{Default Program} , and change \textcolor{normalGreen}{it} .\\
			\hline
	\end{tabularx}}
	\caption{Translation example from the eu-en test set.}\label{example}
	\vspace{-2pt}
\end{table}

\begin{figure*}[t!]
	\centering
	\includegraphics[width=\linewidth]{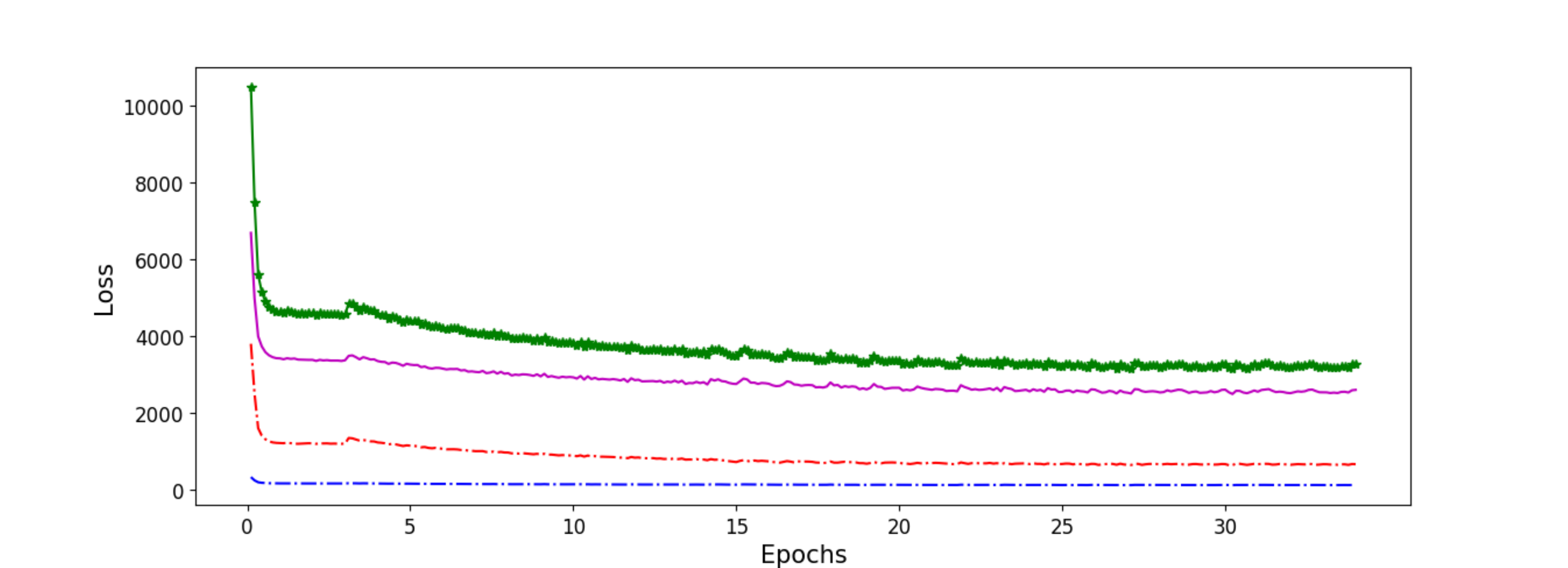}
	\caption{Plot of the values of various loss functions during training of our model over the en-fr training set: 
		\textcolor{green}{\textbf{green}}: training loss (NLL + ($\lambda = 20$) ReWE (CEL); Eq.\ref{eq:combined_loss}); 
		\textcolor{red}{\textbf{red}}: NLL loss; 
		\textcolor{blue}{\textbf{blue}}: ReWE (CEL) loss; 
		\textcolor{magenta}{\textbf{magenta}}: ReWE (CEL) loss scaled by $\lambda = 20$. 
		Each point in the graph is an average value of the corresponding loss over 25,000 sentences.}
	\label{fig:loss}
	\vspace{-8pt}
\end{figure*}

\subsection{Qualitative comparison}
\label{suc_sec_qr}

To further explore the improvements obtained with ReWE, we have qualitatively compared several translations provided by the baseline and the baseline + ReWE (CEL), trained with identical seeds. Overall, we have noted a number of instances where ReWE has provided translations with more information from the source (higher adequacy). For reasons of space, we report only one example in Table \ref{example}, but more examples are available in the supplementary material. In the example, the baseline has chosen a generic word, ``program'', while ReWE has been capable of correctly predicting ``Default Program'' and being specific about the object, ``it''.

\subsection{Discussion}
\label{sub_sec:discussion}

%

To further explore the behaviour of the ReWE loss, Figure \ref{fig:loss} plots the values of the NLL and ReWE (CEL) losses during training of our model over the en-fr training set.  The natural values of the ReWE (CEL) loss (\textcolor{blue}{\textbf{blue}} curve) are much lower than those of the NLL loss (\textcolor{red}{\textbf{red}} curve), and thus its contribution to the gradient is likely to be limited. However, when scaled up by a factor of $\lambda = 20$ (\textcolor{magenta}{\textbf{magenta}} curve), its influence on the gradient becomes more marked. Empirically, both the NLL and ReWE (CEL) losses decrease as the training progresses and the total loss (\textcolor{green}{\textbf{green}} curve) decreases. As shown in the results, this combined training objective has been able to lead to improved translation results.

Conversely, the MSE loss has not exhibited a similarly smooth behaviour (supplementary material). Even when brought to scale with the NLL loss, it shows much larger fluctuations as the training progresses. In particular, it shows major increases at the re-starts of the optimizer for the simulated annealing that are not compensated for by the rest of the training. It is easy to speculate that the MSE loss is much more sensitive than the cosine distance to the changes in the weights caused by dropout and the re-starts. As such, it seems less suited for use as training objective.

\section{Conclusion}
\label{sec:Conclusion}

In this paper, we have proposed a new regularization technique for NMT (ReWE) based on a joint learning setting in which a seq2seq model simultaneously learns to a) predict the next word in the translation and b) regress toward its word embedding. The results over three parallel corpora have shown that ReWE has consistently improved over both its baseline and recent state-of-the-art results from the literature. As future work, we plan to extend our experiments to better understand the potential of the proposed regularizer, in particular for unsupervised NMT~\cite{artetxe2017unsupervised,lample2018phrase}.

\section{Acknowledgments}
\label{sec:Acknowledgments}

We would like to acknowledge the financial support received from the Capital Markets Cooperative Research Centre (CMCRC), an industry-led research initiative of the Australian Government. We would also like to thank Ben Hachey, Michael Nolan and Nadia Shnier
for their careful reading of our paper and their insightful comments. Finally, we are grateful to the anonymous reviewers for all their comments and suggestions.

\balance

\bibliography{naaclhlt2019}
\bibliographystyle{acl_natbib}

%
%


\appendix

\title{\Large{\bf{Supplementary Material - ReWE: Regressing Word Embeddings\\for Regularization of Neural Machine Translation Systems}}}
\author{}

\maketitle
\thispagestyle{empty}

\section{Training and hyperparameters}
\label{sec:supplemental}
In this appendix we provide all the information required to reproduce our results. The models have been implemented by modifying OpenNMT \cite{klein2017opennm} and we will release our code publicly immediately after the anonymity period. All the code is already available to the reviewers as supplementary material. 

To build a strong and current baseline, we have closely followed the indications of \cite{denkowski2017stronger}. The baseline uses a single-layer bidirectional LSTM and a unidirectional LSTM as encoder and decoder, respectively. The attention mechanism is that of~\cite{bahdanau2014}. We have set the size of the LSTMs' hidden layer to 1024, the size of the attention layer to the same size, and the size of the word embeddings to 300. We have initialized the word embeddings with the publicly-available pre-trained vectors from fastText\textsuperscript{6} for each language. The maximum length of the training sentences has been set to 100 tokens. The model vocabulary has been limited to $50,000$ words for both the source and target languages. Words that are not present in the vocabulary are mapped to an $unk$ token, but are later replaced with the corresponding source word with highest attention, following~\cite{luon2015addressing}. For inference, we have used beam search with a beam size of $5$.

We have added ReWE to this baseline, keeping all the aforementioned values unchanged. As mentioned in the paper, ReWE is a stack of two linear layers with a ReLU in between. The first linear layer reduces vector $s_j$ from size $1024$ to $200$. After the ReLU, the second linear layer expands the vector from size $200$ to $300$, which is the size of the word embeddings. The value for $\lambda$ has been selected by evaluating the model over the en-fr validation set (see Section 4.2 in the paper).

All the models have been trained until convergence of the perplexity, using the Adam optimizer~\cite{kingma2014adams}, with a maximum step size of $0.0002$, multiple restarts, and learning rate annealing~\cite{denkowski2017stronger}. After three consecutive validation evaluations without perplexity improvement, we halve the learning rate, and we repeate this process $5$ times. After the $5$-th halving, we stop the training if there is no perplexity improvement over $20$ consecutive runs. The batch size is $40$ and the model is evalauted every $25,000$ sentences. 

We have also trained the models at sub-word level using byte pair encoding (BPE)~\cite{sennrich2016neura}. We have learned the sub-word models using the concatenated training sets of all datasets, setting the number of merge operations to $32,000$ for en-fr and cs-en, and to $8,000$ for eu-en, given its much smaller size. We have also pre-trained word embeddings of size $300$ for the new sub-word vocabularies, and used them for initialization of the word embeddings.

For each model, we have reported the average BLEU score~\cite{papineni2002bleus} of $10$ independent runs, except for the selection of $\lambda$ where we have averaged only $3$ independent runs. \blfootnote{\textsuperscript{6} https://fasttext.cc/docs/en/crawl-vectors.html}

\section{Translation examples}
\label{sec:trans_examples}

In this section we showcase more examples of translations made by the model with and without ReWE for all the language pairs evaluated in the paper (en-fr, cs-en and eu-en). In general the translations made by ReWE seem to preserve a higher amount of information from the original source sentence, which is often referred to as higher ``adequacy''.

\begin{table*}[t]
	\centering
	\resizebox{0.7\textwidth}{!}{\begin{tabularx}{2.1\columnwidth}{|l X|}
			
			\hline
			\textbf{Src}: &Even in just the past few years , we've greatly expanded our knowledge of how Earth fits within \textcolor{normalGreen}{the context} of our universe .\\
			\textbf{Ref}: &Rien qu' au cours des derni\`{e}res ann\'{e}es , nous avons beaucoup appris sur la fa\c{c}on dont la Terre s' int\`{e}gre dans \textcolor{normalGreen}{le contexte} de notre univers .\\
			\textbf{Baseline}: &M\^{e}me ces derni\`{e}res ann\'{e}es , nous avons \'{e}norm\'{e}ment \'{e}largi notre connaissance de la mani\`{e}re dont la Terre s' adapte au sein \textcolor{red}{de notre univers} .\\
			\textbf{Baseline+ReWE}: &M\^{e}me ces derni\`{e}res ann\'{e}es , nous avons grandement \'{e}largi nos connaissances sur la mani\`{e}re dont la Terre s' adapte dans \textcolor{normalGreen}{le contexte} de notre univers .\\
			\hline
			\textbf{Src}: &\textcolor{normalGreen}{So ,} the first example is `` a long time ago . ''\\
			\textbf{Ref}: &\textcolor{normalGreen}{Donc ,} le premier exemple est `` il y a longtemps '' .\\
			\textbf{Baseline}: &\textcolor{red}{Le premier exemple} est `` il y a longtemps . ''\\
			\textbf{Baseline+ReWE}: &\textcolor{normalGreen}{Donc ,} le premier exemple est `` il y a longtemps . ''\\
			\hline
			\textbf{Src}: &And let me tell you , kids with \textcolor{normalGreen}{power tools} are awesome and safe .\\
			\textbf{Ref}: &Laissez-moi vous dire que les enfants sont g\'{e}niaux et prudents avec des \textcolor{normalGreen}{outils \'{e}lectriques} .\\
			\textbf{Baseline}: &Et laissez moi vous dire , les enfants avec les \textcolor{red}{outils du pouvoir} sont stup\'{e}fiantes et s\^{u}rs .\\
			\textbf{Baseline+ReWE}: &Laissez-moi vous dire que les enfants avec des \textcolor{normalGreen}{outils \'{e}lectriques} sont stup\'{e}fiantes et s\^{u}rs .\\
			\hline
	\end{tabularx}}
	\caption{Translation examples from en-fr test set.}\label{example_fr}
\end{table*}

\begin{table*}[t]
	\centering
	\resizebox{0.7\textwidth}{!}{\begin{tabularx}{2.1\columnwidth}{|l X|}
			
			\hline
			\textbf{Src}: &Nikdy toti\v{z} na architekturu neexistovala dobr\'{a} zp\v{e}tn\'{a} vazba .\\
			\textbf{Ref}: &That's because \textcolor{normalGreen}{there's never been} a good feedback loop in architecture .\\
			\textbf{Baseline}: &\textcolor{red}{You've never had} a good feedback in architecture .\\
			\textbf{Baseline+ReWE}: &\textcolor{normalGreen}{It's never been} a good feedback in architecture .\\
			\hline
			\textbf{Src}: &P\v{r}ed tis\'{i}ci lety jste se museli proj\'{i}t do vedlej\v{s}\'{i} vesnice , abyste se na n\v{e}jakou budovu pod\'{i}vali .\\
			\textbf{Ref}: &A thousand years ago , you would have had to have walked \textcolor{normalGreen}{to the village next door} to see a building .\\
			\textbf{Baseline}: &A thousand years ago , you had to go \textcolor{red}{to the side of the village} to look at some building .\\
			\textbf{Baseline+ReWE}: &A thousand years ago , you had \textcolor{normalGreen}{to go to the next village} to look at some building .\\
			\hline
			\textbf{Src}: &V tomto okam\v{z}iku se v\'{a}m uvnit\v{r} hlavy prom\'{i}t\'{a} film.\\
			\textbf{Ref}: &Right now you have a movie playing \textcolor{normalGreen}{inside your head} .\\
			\textbf{Baseline}: &And at that point , I 'm going to give you a film \textcolor{red}{inside a film} .\\
			\textbf{Baseline+ReWE}: &In this point , you have a film \textcolor{normalGreen}{inside the head} .\\
			\hline
	\end{tabularx}}
	\caption{Translation examples from cs-en test set.}\label{example_cs}
\end{table*}

\begin{table*}[!htbp]
	\centering
	\resizebox{0.7\textwidth}{!}{\begin{tabularx}{2.1\columnwidth}{|l X|}
			
			\hline
			\textbf{Src}: &Hautatu Kontrol panela $\rightarrow$ Programa lehenetsiak , eta aldatu bertan .\\
			\textbf{Ref}: &Go to Control Panel $\rightarrow$ \textcolor{normalGreen}{Default programs} , and change it there .\\
			\textbf{Baseline}: &Select the Control Panel $\rightarrow$ \textcolor{red}{program} , and change .\\
			\textbf{Baseline+ReWE}: &Select the Control Panel $\rightarrow$ \textcolor{normalGreen}{Default Program} , and change it .\\
			\hline
			\textbf{Src}: &Hautatu Diapositiba aukerak $\rightarrow$ Pantaila $\rightarrow$ Erakutsi ataza barra . Aukeratu ireki nahi duzun programa . Sakatu PowerPoint ikonoa aurkezpenera itzultzeko .\\
			\textbf{Ref}: &Select the Slide Options $\rightarrow$ Screen $\rightarrow$ Show Taskbar . Choose a program you 'd like to open . Click the \textcolor{normalGreen}{PowerPoint} icon to return to the presentation .\\
			\textbf{Baseline}: &Select the Slide Options $\rightarrow$ Display the Show tasbar . Choose the program you want to open . Click \textcolor{red}{the program} to return the presentation to the presentation .\\
			\textbf{Baseline+ReWE}: &Select the Slide Options $\rightarrow$ Display $\rightarrow$ Show Screen Bar . Choose the program that you want to open . Press \textcolor{normalGreen}{PowerPoint} icon to return to the presentation .\\
			\hline
			\textbf{Src}: &Konektatu gailua \textcolor{normalGreen}{energia iturri} batera . Sakatu Ezarpenak $\rightarrow$ Orokorra $\rightarrow$ Software eguneratzea . Sakatu Deskargatu eta instalatu . Sakatu Instalatu deskarga osatzean .\\
			\textbf{Ref}: &Plug in your device to a \textcolor{normalGreen}{power source} . Tap Settings $\rightarrow$ General $\rightarrow$ Software Update . Tap Download and Install . \textcolor{normalGreen}{Tap Install when the download completes .}\\
			\textbf{Baseline}: &Connect the device to the \textcolor{red}{power} . Tap Settings $\rightarrow$ General $\rightarrow$ Software update . Tap Download and install . \textcolor{red}{Click Install to download .}\\
			\textbf{Baseline+ReWE}: &Connect the device to a \textcolor{normalGreen}{power source} . Tap Settings $\rightarrow$ General $\rightarrow$ Software update . Tap Download and install it . \textcolor{normalGreen}{Click Install when completed Download .}\\
			\hline
	\end{tabularx}}
	\caption{Translation examples from eu-en test set.}\label{example_eu}
\end{table*}

\newpage

\begin{figure*}[t!]
	\centering
	\includegraphics[width=\linewidth]{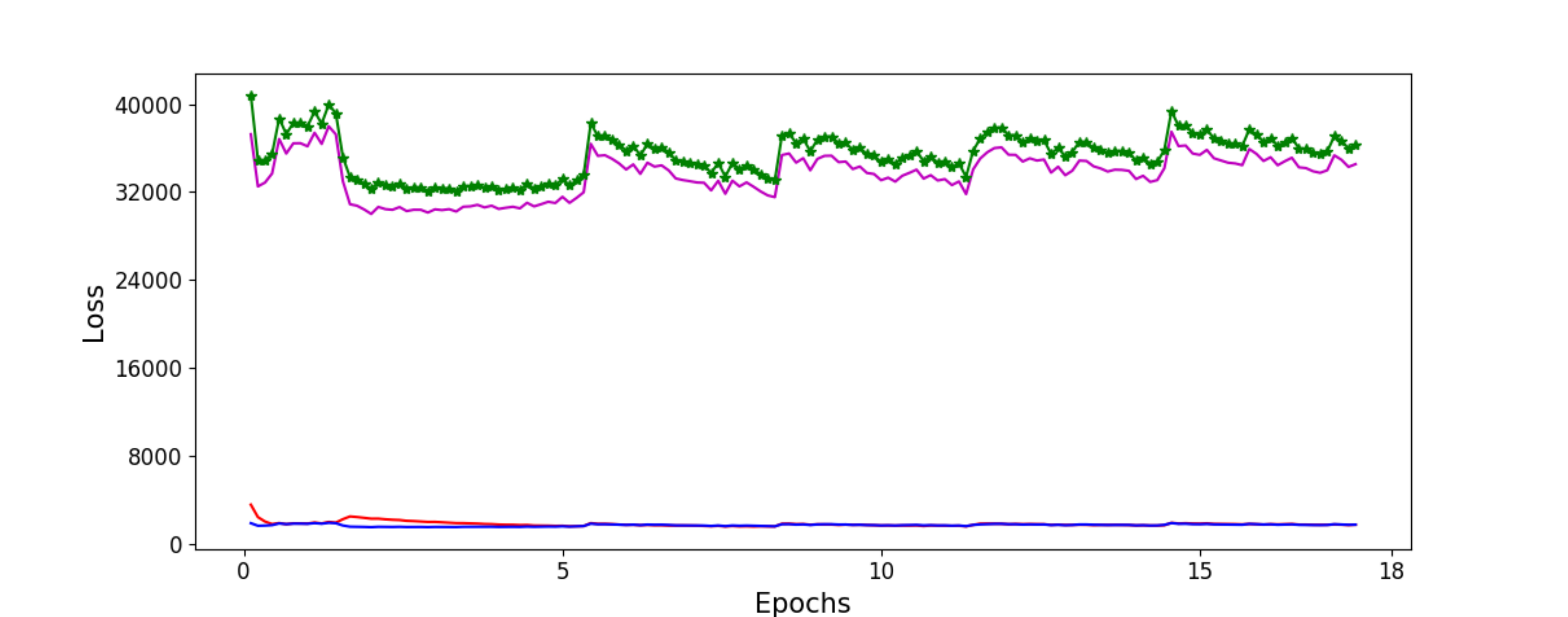}
	\caption{Plot of the values of various loss functions during training of our model over the en-fr training set: 
		\textcolor{green}{\textbf{green}}: training loss (NLL + ($\lambda = 20$) ReWE (MSE); Eq.7); 
		\textcolor{red}{\textbf{red}}: NLL loss; 
		\textcolor{blue}{\textbf{blue}}: ReWE (MSE) loss; 
		\textcolor{magenta}{\textbf{magenta}}: ReWE (MSE) loss scaled by $\lambda = 20$. 
		Each point in the graph is an average value of the corresponding loss over 25,000 sentences.}
	\label{fig:loss}
	\vspace{-8pt}
\end{figure*}

\section{Constrastive experiments}

To gain further insight on the performance of the proposed technique, we have added two contrastive experiments. The first one (Contrastive A) removes ReWE from the architecture, but still retains the combined loss function (Eq. 7 in the paper). Instead of computing the $ReWE_{loss}$ between the ground-truth embedding and the regressed embedding, we compute it between the ground-truth embedding and the word embedding of the predicted word, $e(\arg\!\max p_j)$. This experiment probes whether the system can leverage the distributional properties of the word embeddings without explicitly predicting them.

\begin{table}[t]
	\centering
	\resizebox{0.25\textwidth}{!}{\begin{tabularx}{0.6\columnwidth}{|l|l X|}
			
			\hline
			\textbf{Dataset}&\multicolumn{2}{c|}{\textbf{BLEU}}\\
			&Word&BPE\\
			\hline
			en-fr&$33.82$&$33.37$\\
			cs-en&$20.70$&$22.53$\\
			eu-en&$12.15$&$17.53$\\
			\hline
	\end{tabularx}}
	\caption{Results of the Contrastive A experiment ($\lambda=0.2$; average of $10$ models trained independently from different random seeds).}\label{tab:results_contrastive}
	\vspace{-6pt}	
\end{table}

The second contrastive experiment (Contrastive B) relies solely on ReWE for both training and inference. Instead of the combined loss function, we only use the $ReWE_{loss}$ for training. At inference time, a search is performed over the embedding space to find the nearest neighbor of the predicted embedding and use it as the predicted word. This experiment aims to explore whether the word embeddings can completely replace the usual categorical prediction.

Table \ref{tab:results_contrastive} shows the results for the Contrastive A experiment. For this experiment, the value of $\lambda$ has been specifically tuned over the er-fr validation set (highest score for $\lambda = 0.2$). However, this configuration has rarely improved over our baseline (e.g., on the eu-en dataset), and it has performed considerably worse with the en-fr pair. This shows that, in comparison, the proposed joint learning is a much more effective setting.

In turn, the Contrastive B experiment has achieved much lower BLEU scores. The first experiment over the cs-en dataset reported only 12.71 BLEU points (average of $10$ independent runs), approximately half of the other models. Due to this poor result, we have not carried out this experiment further. Our interpretation of this result is that targeting the word embedding is an effective regularizer in the continuous domain, but the conversion of the predicted word embedding to a categorical value is prone to errors from closer neighbors.

\section{Behaviour of the ReWE (MSE) loss}

Figure \ref{fig:loss} plots the values of the NLL and ReWE (MSE) losses during training of our model over the en-fr training set. The ReWE (MSE) loss shows large fluctuations as the training progresses, with major increases at the re-starts of the optimizer for the simulated annealing
that are not compensated for by the rest of the training.

\balance

\end{document}